\documentclass{article}

\usepackage{microtype}
\usepackage{graphicx}
\usepackage{subfigure}
\usepackage{booktabs} % for professional tables
\usepackage{wrapfig}
\usepackage{times}
\usepackage{epsfig}
\usepackage{amsmath}
\usepackage{amssymb}
\usepackage{pifont}

\usepackage{multirow}
\usepackage{tabularx}
\usepackage{dsfont}
\usepackage{url}
\usepackage{subfigure}
\usepackage{amsthm}
\usepackage[export]{adjustbox}

\theoremstyle{definition}

\theoremstyle{remark}

\usepackage{hyperref}

\usepackage[accepted]{icml2020}

\icmltitlerunning{ Towards a Hypothesis on Visual Transformation based Self-Supervision  }

\begin{document}

\twocolumn[
\icmltitle{ Towards a Hypothesis on Visual Transformation based Self-Supervision  }




\begin{icmlauthorlist}
\icmlauthor{Dipan K. Pal}{to}
\icmlauthor{Sreena Nallamothu}{to}
\icmlauthor{Marios Savvides}{to}
%\icmlauthor{Iaesut Saoeu}{ed}
%\icmlauthor{Fiuea Rrrr}{to}
%\icmlauthor{Tateu H.~Yasehe}{ed,to,goo}
%\icmlauthor{Aaoeu Iasoh}{goo}
%\icmlauthor{Buiui Eueu}{ed}
%\icmlauthor{Aeuia Zzzz}{ed}
%\icmlauthor{Bieea C.~Yyyy}{to,goo}
%\icmlauthor{Teoau Xxxx}{ed}
%\icmlauthor{Eee Pppp}{ed}
\end{icmlauthorlist}

\icmlaffiliation{to}{Department of Electrical and Computer Engg., Carnegie Mellon University, Pittsburgh, USA}
%\icmlaffiliation{goo}{Googol ShallowMind, New London, Michigan, USA}
%\icmlaffiliation{ed}{School of Computation, University of Edenborrow, Edenborrow, United Kingdom}

\icmlcorrespondingauthor{Dipan K. Pal}{dipanp@andrew.cmu.edu}
%\icmlcorrespondingauthor{Eee Pppp}{ep@eden.co.uk}

\icmlkeywords{Machine Learning, ICML}

\vskip 0.3in
]



\printAffiliationsAndNotice{}  % leave blank if no need to mention equal contribution

\begin{abstract}
We propose the first qualitative hypothesis characterizing the behavior of visual transformation based self-supervision, called the VTSS hypothesis. Given a dataset upon which a self-supervised task is performed while predicting instantiations of a transformation, the hypothesis states that if the predicted instantiations of the transformations are already present in the dataset, then the representation learned will be less useful. The hypothesis was derived by observing a key constraint in the application of self-supervision using a particular transformation. This constraint, which we term the transformation conflict for this paper, forces a network to learn degenerative features thereby reducing the usefulness of the representation. The VTSS hypothesis helps us identify transformations that have the potential to be effective as a self-supervision task. Further, it helps to generally predict whether a particular transformation based self-supervision technique would be effective or not for a particular dataset. We provide extensive evaluations on CIFAR 10, CIFAR 100, SVHN and FMNIST confirming the hypothesis and the trends it predicts. We also propose  novel cost-effective self-supervision techniques based on translation and scale, which when combined with rotation outperform all transformations applied individually. Overall, the aim of this paper is to shed light on the phenomenon of visual transformation based self-supervision.
\end{abstract}

%%%%%%%%% BODY TEXT
\section{Introduction}

%TODO

%relate to data augmentation

%\textbf{The Problem of Representation Learning.} One of the most fundamental problems in machine learning is problem of representation learning. There have been a plethora of efforts and approaches towards this problem. However, it has become apparent that future endeavours would greatly benefit from the focus on unsupervised learning. This is due a variety of reasons not limited to the growing need for more data, coupled with the steady high cost of labelling data and finally channeled by an inherent interest towards imitating or understanding biological learning. Among these, a sub-class of unsupervised learning has emerged which has proven to be a worthy area for persistent exploration, called self-supervised learning. 

\textbf{The Mystery of Self-Supervision.} Self-supervision loosely refers to the class of representation learning techniques, where it is cost effective to produce effective supervision for models using some function of the data itself. Indeed in many cases, the data becomes its own ground-truth. While a lot of efforts are being directed towards developing more effective techniques \cite{hendrycks2019using,sermanet2018time,zhang2019aet,trinh2019selfie,doersch2015unsupervised}, there has not been enough attention on the problem of understanding these techniques or at least a sub-set of them at a more fundamental level. Indeed while there have been many efforts which introduced self-supervision in different forms \cite{zhang2016colorful,zhai2019visual,gidaris2018unsupervised}, there have been only a few efforts which shed more light into related phenomenon \cite{goyal2019scaling}. In one such work, the authors focus on the trends (and the lack of) that different architecture choices have on the performance of the learnt representations \cite{kolesnikov2019revisiting}. One emerging technique that has proven to learn useful representations while being deceptively elementary is the study of RotNet \cite{gidaris2018unsupervised,feng2019self}. RotNet takes an input image and applies specific rotations to it. The network is then tasked with predicting the correct rotation applied. In doing so, it was shown to learn useful representations which could be used in many diverse downstream applications. It was argued that however, that in learning to predict rotations, the network is forced to attend to the shapes, sizes and relative positions of visual concepts in the image. This explanation though intuitive, does not help us to better understand the behavior of the technique. For instance, it does not help us answer the question: \textit{Under what conditions would a particular method succeed in learning useful representations?}

\begin{figure}%{r}{0.5\textwidth}
    \begin{center}
        \includegraphics[width=0.95\columnwidth,valign=m]{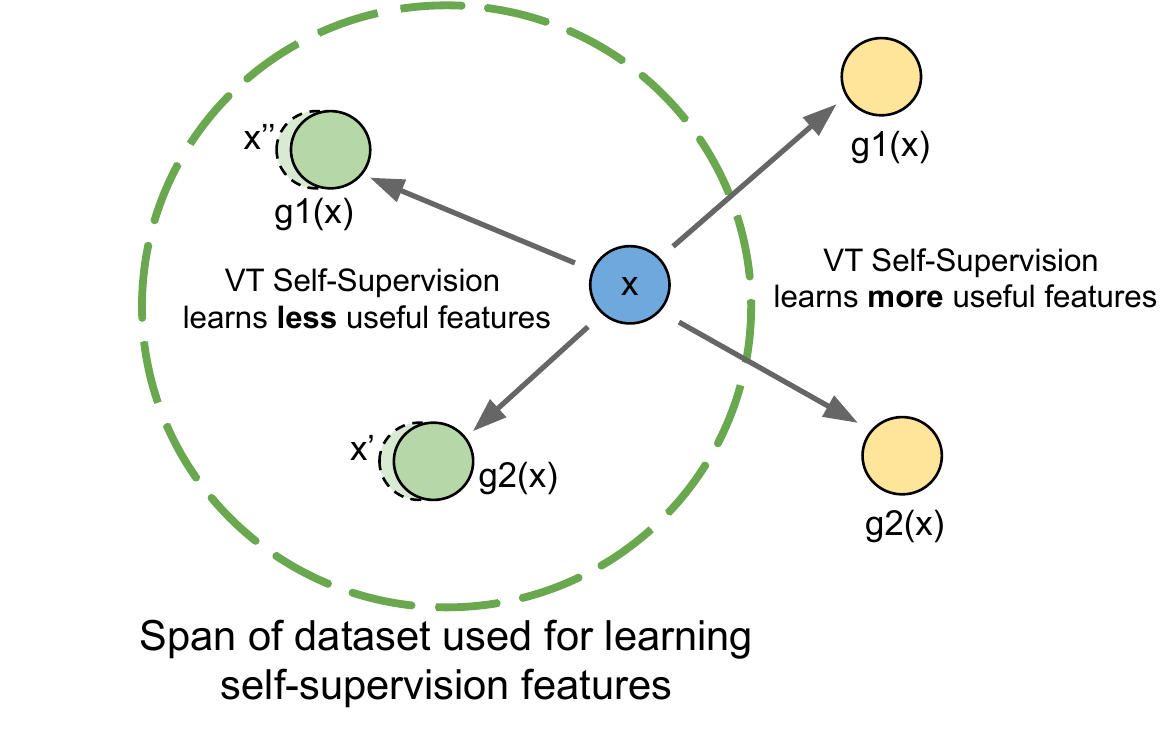}
    \end{center}
    %\vspace{-0.5cm}
\caption{ \textbf{The Visual Transformation Self-Supervision (VTSS) Hypothesis:} Given a dataset upon which a self-supervised task is performed while predicting instantiations of a transformation, the hypothesis states that if the predicted instantiations of the transformations are already present in the dataset, then the representation learned will be less useful. }
\label{fig_VTSS}
    \vspace{-0.5cm}%

\end{figure}

\textbf{Towards an Initial Hypothesis of Self-Supervision.} This study hopes to provide initial answers or directions to such questions. We first model supervised classification within a transformation framework. We then define a general self-supervision task utilizing a particular transformation within the same framework. In doing so, we discover an  effect depending on the dataset and the transformation used, which would force a network to learn noisy or degenerate features. Finding a simple condition which avoids this constraint, we arrive at a visual transformation based self-supervision
(VTSS) hypothesis (see Fig.~\ref{fig_VTSS}). Consider a dataset that is to be used to learn features through a pretext VTSS task of augmenting the data with instantiations of transformation $g$  and then predicting it (the dataset is indicated by the large dotted green circle in Fig.~\ref{fig_VTSS}). Let $x$ be a sample in the dataset, with $g_1(x)$ and $g_2(x)$ being two instantiations of the  transformation $g$ applied to $x$ (solid green dots)  as part of the pretext VTSS task. The VTSS hypothesis predicts that for any $x$, if $g_1(x)$ and $g_2(x)$ (solid green dots) lie  close to samples already in the dataset (light green dotted samples), then the features learnt using the VTSS task of predicting $g$ will learn \textit{less} useful features \emph{i.e.} the features will be less suited for the main task such as downstream classification using that representation. On the other hand, if the augmented samples $g_1(x)$ and $g_2(x)$  are not present in the dataset and map to points outside the dataset (solid yellow dots), then the features learned will be \textit{more} useful. The hypothesis is based off the observation that for a given VTSS task based on $g$, if $g(x)$ is close in distance to a sample $x'$ already in the dataset or exactly the same (\emph{i.e.} $g_1(x)=x'$ and $g_2(x)=x''$, dotted green dots), then this creates a destructive effect called the \textbf{transformation conflict}. This effect forces a network to learn degenerate features. In the figure, VTSS will learn more degenerate features if $g(x)$ maps to the green dots rather than the yellow dots.

This hypothesis is the first attempt to characterize and move towards a theory of the behavior of self-supervision techniques utilizing visual transformations. The VTSS hypothesis also offers practical applications. For instance, the hypothesis can predict or suggest reasonable transformations to use as a prediction task in VTSS given a particular dataset. Indeed, we introduce two novel VTSS tasks based on translation and scale respectively which we study in our experiments. The hypothesis can also help predict the relative trends in performance between VTSS tasks based on one or more transformations on a particular dataset based on some of the dataset properties. We confirm the VTSS hypothesis and a few trends that it predicts in our experiments.

\textbf{Our Contributions.} 1) We provide an initial step towards a theory of behavior of visual transformation based self-supervision (VTSS) techniques in the form of a qualitative hypothesis. The hypothesis describes a condition when a self-supervision technique based on a particular visual transformation would succeed. 2) We use the hypothesis to propose two novel self-supervision tasks based on translation and scale and argue why they might be effective in particular cases. 3) We provide extensive empirical evaluation on CIFAR 10, CIFAR 100, SVHN and FMNIST providing empirical evidence towards confirming the hypothesis for all transformations studied, \emph{i.e.} translation, rotation and scale. We further propose to combine transformations within the same self-supervision tasks leading to performance gains over individual transformations. 4) Finally, we provide an array of ablation studies and observations on VTSS using rotation, translation and scale. For instance, we find that improvements in semi-supervised classification performance provided by unlabelled data used for self-supervision is very similar to that provided by having the same amount of labelled data used for semi-supervised training.

%\cite{hendrycks2019using} The authors explore some robustness properties learnt from self-supervised representations which were learnt using rotations. They argue that predicting rotations requires understanding object shape. However, this effect can only occur if the transformation does not exist in the dataset which is hyp 1.

%%%%%%%%%%%%%%%%%%%%%%%%%%%%%%%%%%%%%%%%%%%%%%%%%%%%%%%%%%%%%%%%%%%%%%%%%%%%%%%%%%%%%%%%%%%%%%%%%%%%%%%%%%%%%%%%%%%%%%%%%%%%%%%%%%%%%%%%%%%%%%%%%
%\section{Prior Art in Self-Supervision}
%There have been a plethora of works of self-supervision. Works based on egomotion \cite{agrawal2015learning}, predicting  

%\textbf{Metric based Self Supervision}

%Time Constrastive Networks attempt to learn a viewpoint invariant representation \cite{sermanet2018time} by having representations from different viewpoints however simultaneous in time, close in a metric space. 

%The following papers use spatial patch location prediction in some way \cite{trinh2019selfie,doersch2015unsupervised}

%\cite{zhang2019aet} predicts transformations from \textit{representations} and not bare images.

%%%%%%%%%%%%%%%%%%%%%%%%%%%%%%%%%%%%%%%%%%%%%%%%%%%%%%%%%%%%%%%%%%%%%%%%%%%%%%%%%%%%%%%%%%%%%%%%%%%%%%%%%%%%%%%%%%%%%%%%%%%%%%%%%%%%%%%%%%%%%%%%%

\section{Prior Art}

Self-supervised learning has recently garnered a lot of attention from the community. For a brief overview of various techniques and methods, we encourage the reader to refer to \cite{jing2019self}.   Self-supervision has proven to be effective in areas other than vision such as NLP \cite{devlin2018bert,wu2019self}, robotics and reinforcement learning \cite{pathak2019self,jang2018grasp2vec,ebert2018robustness,nair2017combining,murali2018cassl}. One of the main methods of performing self-supervision to learning useful features is to solve a pretext task. Such a task is chosen that is ideally computationally cheap and more importantly, one that allows the training on a `good' representation. 
\cite{he2019momentum}

\textbf{Pretext task based methods.} Several pretext tasks have been proposed for self-supervision. For instance, solving a patch-based jigsaw puzzle \cite{doersch2015unsupervised,noroozi2016unsupervised,kim2018learning}, predicting color channels  \cite{zhang2016colorful,larsson2016learning,zhang2017split}, predicting rotations on images \cite{gidaris2018unsupervised}, learning features through reversing inpainting \cite{pathak2016context}, and learning to count \cite{noroozi2017representation}. Utilizing spatial context as a supervision signal was also explored \cite{Doersch_2015_ICCV}. Studies found that learning robustness to corruptions in input was also an effective pretext task \cite{pathak2016context,vincent2008extracting}. Geometric transformations were found to be useful to learn representations in the study \cite{dosovitskiy2014discriminative}, however it did not predict the instance or the transformation but rather aimed to learn invariance towards them. This is different from a VTSS task which predicts the exact instance of the transformation applied. Another recent task that has shown considerable promise is  to match a query representation to other keys in a set belonging to the same image \cite{he2019momentum}. Other methods utilize clustering even after utilizing a pretext task \cite{noroozi2018boosting,caron2018deep}.    There also have been similar pretext tasks proposed on videos such as solving jigsaws on video frames \cite{ahsan2019video,wei2019iterative}.  Augmenting and then predicting rotations on videos was also found to offer a useful self-supervision signal \cite{jing2018self}. Contrastive predictive coding \cite{oord2018representation} and contrastive multiview coding \cite{tian2019contrastive} are other successful  method which utilized some form of prediction of the data.  In the real-world, the laws of physics along with time constantly provide valuable transforming data. These temporal based visual transformations can be yet another source of supervision as explored by \cite{misra2016shuffle,wei2018learning,lee2017unsupervised,pathak2017learning}.

%%%%%%%%%%%%%%%%%%%%%%%%%%%%%%%%%%%%%%%%%%%%%%%%%%%%%%%%%%%%%%%%%%%%%%%%%%%%%%%%%%%%%%%%%%%%%%%%%%%%%%%%%%%%%%%%%%%%%%%%%%%%%%%%%%%%%%%%%%%%%%%%%
\section{Supervision from the Transformation Perspective}

\textbf{The Transformation Model of Visual Images.} We adopt a model of data transformations which accounts for all the variation that is seen in general data. Given an image which when vectorized results in a seed vector $x$ sampled from a seed distribution $P_x$, it is first  acted upon by the transformation $g_k: \mathbb{R}^d \rightarrow \mathbb{R}^d$ \emph{i.e.} $g_k(x;~\theta_k)$. This transformation $g_k \mathcal{G}$ is parameterized by $\theta_k$ and generates a sample from the specific class $k$.   For brevity, we drop the notation for the parameters and express a sample as $g_k( x)$.  The transformations $g_k$ are complex and non-linear, and can introduce \textit{features} into a particular sample that to a receiver might appear to be associated to a particular class. For instance, $g_k$ could potentially be features of a particular individual in the case of face recognition, or features of a face at a particular pose for pose estimation.

\textbf{The Transformation Paradigm of Supervised Classification.} In the context of supervised classification, different instantiations of the parameters $\theta_k$ along with different seed vectors $x$ give rise to all of the samples that can be observed for class $k$ in training and testing. Training data is assumed to include only a subset of all possible combinations of the parameters. Testing data would be sampled from the remaining space of combinations. Note that we do not account for any relation or overlap between $g_k$ and $g_{j}$ for classes $k$ and $j$. For classification, given an input image $g_k( x)$, a classifier $F$ is tasked with predicting the class output $k$. In other words, the task is to predict which of the $k$ transformations from the set $\mathcal{G} $ was applied. 

\textbf{Self-Supervision from the Visual Transformation Perspective.} The transformation framework is general and can be applied to any classification problem, including self-supervision. A general self-supervision task utilizing a particular transformation $g$ would allow all data variation or transformations to be accounted for in the seed distribution $P_x$ independent of $g$. In fact, the different classes are simply $g_k$ where $k$ is a \textit{particular } instantiation of the transformation $g$. For instance, self-supervision based on rotation would be modelled as $g$ being the in-plane rotation transformation and $k$ being a particular instantiation of it \emph{e.g.} $90^\circ$ clockwise.  Note that for a self-supervision task under this framework, \textit{all} image data is modelled as seed vectors in the distribution $P_x$ with $g_k$ being a particular instantiation of a transformation, including the identity transformation $e$.

%%%%%%%%%%%%%%%%%%%%%%%%%%%%%%%%%%%%%%%%%%%%%%%%%%%%%%%%%%%%%%%%%%%%%%%%%%%%%%%%%%%%%%%%%%%%%%%%%%%%%%%%%%%%%%%%%%%%%%%%%%%%%%%%%%%%%%%%%%%%%%%%%
%\textbf{Translation and Scale based Self-Supervision} Rotation has been successfully explored as a self-supervision task \cite{gidaris2018unsupervised}. However, only an intuitive explanation was provided as to why rotation was chosen specifically. It was argued that it would be very difficult for the network to solve the rotation task without first learning features regarding objects and its parts. This intuition was supported by visualization of filters and experimental performance gains. Nonetheless, there lacked a discussion as to why other transformations could not be utilized for self-supervision. In this study, we introduce and propose to explore self-supervision tasks based on \textit{translation} and \textit{scale}. Here, $g$ would directly be the translation or scale transformation.  We believe that self-supervision tasks based on translation and scale have not been explored before.

\section{A Hypothesis on   Visual Transformation based Self-Supervision}

 For our purpose, we define \textit{usefulness} of a representation  as the semi-supervised classification test accuracy $C$, of a downstream classifier using that representation on a classification task of higher abstraction. A classification task can be loosely termed to be at a  higher abstraction level if it uses a  more complicated transformation set $\mathcal{G}$, than the one used for the self-supervision task.

\textbf{The VTSS Hypothesis:} Let $\mathcal{G} = \{ g_k \} \cup \{e\}~~ \forall k=1.\textcolor{blue}{..}|\mathcal{G}|$ be a set of transformations acting on a vector space $\mathbb{R}^d$ with $e$ being the identity transformation and further with $g(x')\in \mathbb{R}^d ~~\forall x'\in \mathbb{R}^d ~~\forall g\in \mathcal{G}$. Let $\mathcal{X}$ be the set of \textit{all} seed vectors $x\in \mathbb{R}^d$. Finally, we simulate a dataset  with a set of \textit{pre-existing} transformations $\mathcal{H}$, by letting $\mathcal{H} = \{ h ~~|~~ h(x)\in \mathcal{X} ~~\forall x\in \mathcal{X}\}  $. Now, for a usefulness measure of $C$ of a representation $F(x)$ that is trained using a transformation based self-supervision task which predicts instantiations of $\mathcal{G}$, the VTSS hypothesis predicts
\begin{equation}
    C \varpropto (| \mathcal{G}\cap \mathcal{H} |)^{-1}\label{eq_vtss_hyp}
\end{equation}

In other words, consider a dataset of images and a visual transformation based self-supervision (VTSS) task that predicts instantiations of $\mathcal{G}$. Then if the dataset already contains a lot of variations in its samples due to any of the transformations in $\mathcal{G}$, then the VTSS hypothesis predicts that the features learnt on that dataset using the VTSS task corresponding to $\mathcal{G}$ will \textit{not} produce useful features or the usefulness will be \textit{diminished}. In the hypothesis statement, the transformation set $\mathcal{H}$ is the set of all possible transformations and variations that exist in the dataset $\mathcal{X}$.  Thus, if $\mathcal{G}$ and $\mathcal{H}$ have a lot of transformations in common, $C$ decreases. In other words,  $C$ is inversely proportional to the number of transformations common between $\mathcal{G}$ and $\mathcal{H}$. It is important to note however that every \textit{instantiation} of a transformation is considered different. Therefore, a rotation by $45^\circ$ clockwise is a different transformation than  a rotation by $90^\circ$ clockwise. Each instantiation can be used as a prediction target while constructing the corresponding VTSS task.  

%The VTSS hypothesis serves as an initial step to understanding the behavior of VTSS tasks and their phenomenon. 

\begin{figure}%{r}{0.5\textwidth}
%\begin{wrapfigure}{r}{0.5\textwidth}
    \begin{center}

        \includegraphics[width=0.95\columnwidth,valign=m]{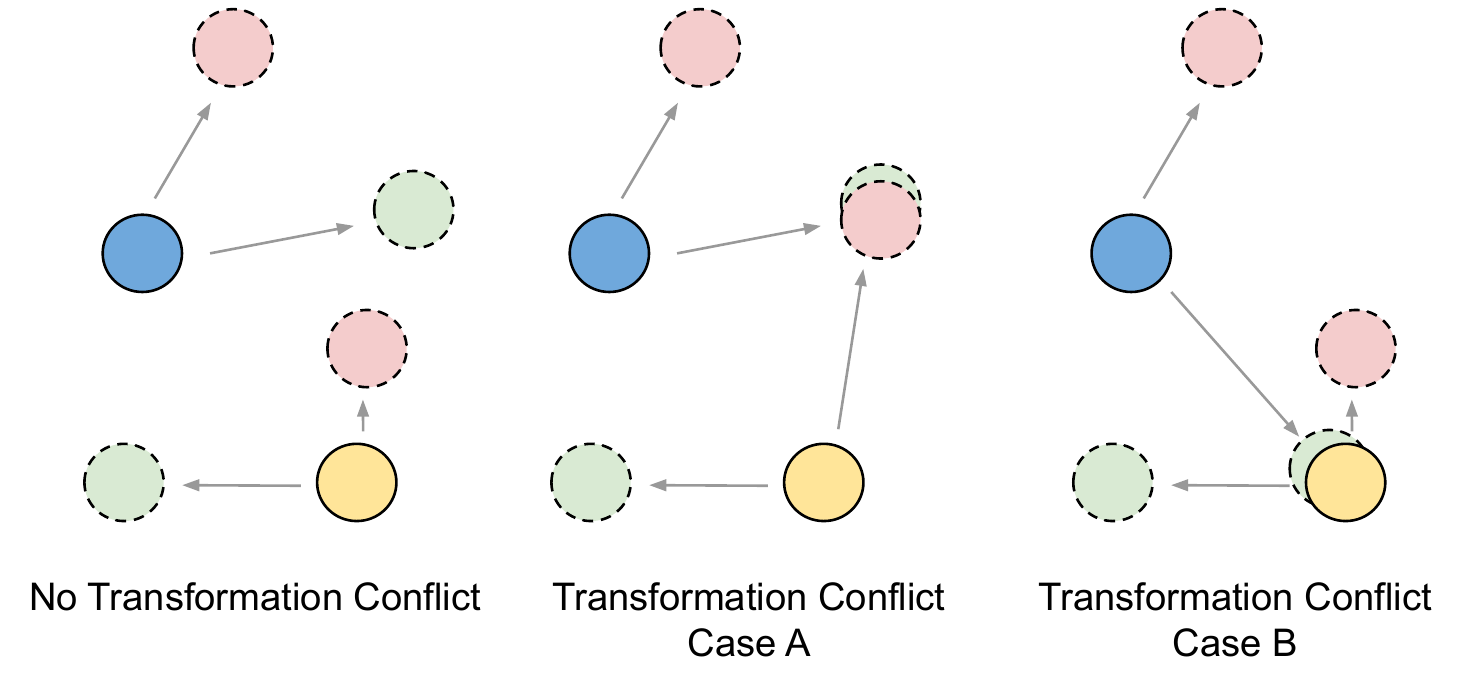}

    \end{center}
    %\vspace{-0.5cm}
\caption{ \textbf{The Effect of Transformation Conflict:} This effect can prevent a self-supervision method from learning a useful representation. There are atleast two potential ways of such an effect manifesting in the dataset. Case A: where two different transformations of two different samples are identical or very similar. Case B: where a transformed version of a sample is identical or very similar to another untransformed sample. }
\label{fig_transformationconflict}
    \vspace{-0.5cm}

\end{figure}
%\end{wrapfigure}

 \textbf{Approaching RotNet from a new perspective.} The hypothesis deters the use of transformations for VTSS tasks which are already present in the data. This might discourage us from utilizing in-plane rotations as a VTSS task since small yet appreciable  amounts of in-plane rotation exist in most real-world datasets. However, we must recall that each \textit{instantiation } of the transformation is considered different. Hence if we consider a rotation angles large enough such as $\{0^\circ, 90^\circ, 180^\circ, 270^\circ\}$, they are unlikely to exist in the dataset.  Thus, the VTSS hypothesis predicts that in-plane rotations would be an effective VTSS task provided the range of rotation is large enough and this indeed has been the observation \cite{gidaris2018unsupervised}. 

\textbf{Identifying Effective Transformations for VTSS. } Following this train of thought, it is natural to ask what other transformations can be used for VTSS? Translation and scale are two relatively simple transformations that are easier to apply (especially translation). It would however be a fair observation to make that both transformations are in fact the most common transformations of variation in real-world visual data, which according to our hypothesis would result in an ineffective learning task. However, owing to manual and automated labelling efforts, there are many datasets in which the visual concept of interest is fairly centered in the image across all samples. This creates an opportunity for the direct use of translation as a computationally inexpensive transformation to apply to self-supervision, while being consistent with the VTSS hypothesis. Scale variation as well can be controlled and accounted for. Nonetheless, in many datasets even when the object is localized, there is relatively more scale variation than translation jitter. As part of this study, we propose the use of both translation and scale as VTSS tasks for use whenever the conditions are favorable.
%\textcolor{red}{May be write one line as in how visual concept of interest being centered facilites use of translation.}

\textbf{Predicting Trends in Relative Performance of VTSS on Datasets. } Currently, a barrage of  self-supervision tasks are applied to a particular dataset as part of a trial and error process towards obtaining a desirable level of performance.  It seems that there exists no heuristic to predict even at minimum a trend of effectiveness. The VTSS hypothesis  provides the an initial heuristic to predict trends in relative performance on any given dataset, given some properties of the dataset. There are at times the possibility of estimating how much variation due to a particular transformation might exist in a given dataset. This might be possible due to control or knowledge of the data collection process, coarse estimation through techniques like PCA or more sophisticated techniques such as disentanglement through an information bottleneck \cite{burgess2018understanding}. In such cases, the VTSS hypothesis can assist in rejecting particular VTSS tasks and prioritize others. For instance, a rotation based VTSS technique is predicted to not be beneficial on a dataset such as Rotated MNIST. Indeed, in our experiments, we observe cases when rotation based self-supervision fails completely.

\textbf{Understanding the VTSS hypothesis. } We discussed a few ways the VTSS hypothesis could be useful. We now provide a qualitative explanation for the same. Consider two samples $x$ and $x'$ belonging to a dataset $\mathcal{X}$. Let there be a network $F$ which will be trained for a VTSS task utilizing the transformation set $\mathcal{G}= \{ g_i ~|~i=\{ 1...k \} \} \cup \{ e\}$ where $e$ is the identity transformation and  $|\mathcal{G}|=k+1$. Therefore, $F(x)$ learns to predict one out of $k+1$ outputs. Specifically for any $x$, given $g_i(x)$ as input (where $g_i$ could be identity), $F(x)$ would need to predict the correct \textit{instantiation} of $\mathcal{G}$ including the identity.  Now, the VTSS hypothesis predicts that as long as $\nexists g'\in \mathcal{G}$ s.t. $g'(x)=x'$ or $g'(x')=x$, the VTSS task will learn useful features.  In other words, as long there exists no transformation instantiation in $\mathcal{G}$ such that $x$ and $x'$ can be related to one another through it, a useful feature will be learned. To see why, we assume  $\exists g_k\in \mathcal{G}$ s.t. $g_k(x)=x'$. Under this assumption, the output of $F(g_k(x))$ should be $k$ \emph{i.e.} $F(g_k(x))=k$. However, we also have $F(x') = F(e(x'))=e$ \emph{i.e.} predicting the identity class since $e\in \mathcal{G}$. Notice that a conflict arises with these two equations, 1) $F(x') = F(e(x'))=e$ and also $F(g_k(x))= F(x')= F(e(x'))=k$. Therefore, for the same input $x'$, the network is expected to output two separate classes. We term this phenomenon as a \textbf{transformation conflict} for this paper (see Fig.~\ref{fig_transformationconflict}), and we observe it in our experiments. This condition over the course of many iterations will learn noisy filters. This is because in practice, there exists small differences between $g_k(x)$ and $x'$. The network will be forced to amplify such differences while trying to minimize the loss, leading to noise being learned as features.

\textbf{The Transformation Conflict:} In Fig.~\ref{fig_transformationconflict}, consider a VTSS task of predicting between three instantiations (including the identity) of a transformation $g$ on a two samples (blue and yellow dots) from a dataset. When $g$ is applied to the samples, it results in the corresponding transformed samples (light red and green dotted dots). Each color signifies the specific label or instantiation of a transformation that the network is tasked with predicting (in the figure there are two colored labels, red and green, with the identity transformation being the sample itself). For instance, RotNet \cite{gidaris2018unsupervised} predicts between 4 angles including $0^\circ$. The self-supervised network will take in as input each transformed or original sample (all dots), and predict the corresponding label (transformation instantiation). \textbf{Left:} In this case, each dataset sample is transformed into points that are distinct and away from other transformed samples or data points. Hence, there is no transformation conflict. The VTSS hypothesis in this case predicts that the features learnt will be useful. \textbf{Center: Case A.} Here, one of the samples (yellow) transforms  into a point (corresponding light red)  close to a transformed version (near by light green) of a separate data sample (blue). This presents a way of incurring transformation conflict within the dataset. For the similar inputs of the closely overlapping dotted red and green dots, the network is expected to predict/output both red and green labels. This causes the network to learn degenerate features, as it is trying to maximize discrimination between the two close-by samples. \textbf{Right: Case B.}  Here, one of the samples (blue) transforms  into a point (corresponding dotted light green)  close to another original data point (yellow). This presents a second way of incurring transformation conflict within the dataset.  For the similar inputs of the closely overlapping yellow and dotted green dots, the network is expected to predict/output both green and identity labels. In both cases A and B, the VTSS hypothesis predicts less useful features learnt by the self-supervision task.

%%%%%%%%%%%%%%%%%%%%%%%%%%%%%%%%%%%%%%%%%%%%%%%%%%%%%%%%%%%%%%%%%%%%%%%%%%%%%%%%%%%%%%%%%%%%%%%%%%%%%%%%%%%%%%%%%%%%%%%%%%%%%%%%%%%%%%%%%%%%%%%%%

%%%%%%%%%%%%%%%%%%%%%%%%%%%%%%%%%%%%%%%%%%%%%
\begin{table}
\small
%\parbox{.5\linewidth}{
\centering
\begin{tabular}{l |  c  | c | c     } % centered columns (4 columns)
\hline %inserts double horizontal lines
\hline
\textbf{Rotation}   &    N-way  &  C10    &  SVHN    \\
\hline
\hline
 %No Rand   &  4  &   88.50    &    91.99   \\
%$0^\circ$, $90^\circ$  &  3  &   87.3    &    58.71   &       &     \\
%$0^\circ$, $90^\circ$, $180^\circ$  &  2  &   87.04    &    81.03   &       &      \\
%\hline
Baseline ($0^\circ$)   &  4  &   $88.50_{(91.99)}$   &    $90.08_{( 86.29)}$  \\
$0^\circ$, $90^\circ$  &  4  &   $88.55 _{(  46.93 )}$  &   $89.88   _{( 43.46 )}$  \\
$0^\circ$, $90^\circ$, $180^\circ$  &  4  &   $87.87 _{(   33.87 )}$  &   $89.28   _{(  33.49)}$    \\
$0^\circ$, $90^\circ$, $180^\circ$, $270^\circ$   &  4  &  $ 10   _{(    25 )}$ &   $7.75  _{(  25 )}$   \\
\hline

\hline %inserts double horizontal lines
\hline
\textbf{Translation}   &   N-way  &  C10   &  SVHN   \\
\hline
\hline
Baseline (C)  &   5     &    $85.13     _{(     60.55 )}$  &   $91.16    _{(   76.98 )}$  \\ 
C, U  &    5    &    $83.10   _{(   34.86 )}$    &    $91.02    _{(  43.17 )}$  \\ 
C, U, R  &    5    &   $82.48_{(   33.16 )}$   &     $90.60  _{(   39.59  )}$ \\ 
C, U, D, L   &    5    &    $81.48   _{(  33.24 )}$    &      $90.62  _{(  36.47 )}$  \\ 
C, U, D, L, R  &   5     &    $79.29    _{(  32.44 )}$    &   $90.16   _{(   37.19 )}$    \\ 

\hline
%Fully Supervised  & 10 &      89.68   &    -   &    92.50   &  -    \\

%\hline %inserts double horizontal lines
%\hline
%\textbf{Scale}   &   \# class  &  Semi (C10)  &   Self  (C10)  &  Semi (SVHN)  &   Self (SVHN)   \\
%\hline
%\hline
%Baseline (Original Scale)  &   2     &   31.16     &    97.15      &     7.76    &    51.24  \\ 
%4 pix Scale up   &    2    &   49.69     &  75.01      &    60.33    & 75.04      \\ 
%\hline
%Fully Supervised  & 10 &   92.15    &    -   &    90.55   &  -    \\
\textbf{FS} 3 blocks & 10 &    $91.35 $   &  $80.83$  \\
\textbf{FS} 4 blocks & 10 &  $91.66$  &  $90.57$    \\
\hline
%inserts single line
\end{tabular}
\caption{\textbf{VTSS Hypothesis Confirmation.}   The column of transformation instantiations on the left were added into the \textit{original data} independent of the additional augmentation by the VTSS task. For each dataset, the number denotes the semi-supervision accuracy following the protocol described while having a N-way transformation prediction that is fixed for Rotation and Translation. FS indicate Fully Supervised. The smaller number in the bracket denotes the N-way (number of transformations) test accuracy during the self-supervision task. The VTSS task remains constant while more transformations are added into the original data. This artificially increases $| \mathcal{G}\cap \mathcal{H} |$ in Eq.~\ref{eq_vtss_hyp}. We see that $C$ decreases inversely, leading to confirmation of Eq.~\ref{eq_vtss_hyp} and the VTSS hypothesis.  }
\label{tab_exp_VTSS}
\vspace{-0.5cm}
\end{table}

\section{Experimental Validation}

Our goal through an extensive experimental validation is threefold. 1) To confirm (or find evidence otherwise to) the VTSS hypothesis for VTSS tasks based on rotation and translation and. 2) To explore the efficacy of solving VTSS tasks with individual transformations and additive combinations. 3) To perform ablation studies on VTSS task based on rotation, translation and scale to help gain insights into effects on semi supervision performance \footnote{We provide these ablation study results and discussions in the supplementary due to space constraints}. For these experiments, we utilize the CIFAR 10, CIFAR 100, FMNIST and SVHN datasets\footnote{We provide additional experimental details in the supplementary.}. Our effort is not to maximize any individual performance metric or achieve state-of-the-art performance on any task, but rather to discover overall trends in behavior leading to deeper insights into the phenomenon of self-supervision through visual transformations.

%We validate our hypothesis through experiments on CIFAR 10 using multiple mainstream algorithms that were previously investigated such as RotNet \cite{gidaris2018unsupervised,hendrycks2019using}, Colorization and Jigsaw. 

\textbf{General Experimental Protocol.} For each transformation and dataset, the overall experimental setup remained unchanged. We follow the training protocol introduced in the RotNet study \cite{gidaris2018unsupervised} where a 4 convolution block backbone network is first trained with a VTSS task based on some transformation (rotation, translation and/or scale). This network is tasked with predicting specific transformation instances following the self-supervision protocol. After training, the network weights are frozen and the feature representations from the second convolution block is utilized for training a downstream convolution classifier which is tasked with the usual supervised classification task \emph{i.e.} predicting class labels for a particular dataset. Our choice of exploring the performance trends of the second block is informed by the original RotNet study where the second conv block exhibited maximum performance for CIFAR 10 \cite{gidaris2018unsupervised}. However, since our focus is on discovering overall trends rather than maximizing individual performance numbers, this choice is inconsequential for our study. Thus, the overall learning setting is semi-supervised learning since part of the pipeline utilized the frozen self-supervised weights. The final semi-supervised test accuracies reported on the test data of each dataset utilized this semi-supervised pipeline. 
%Further details including the network architecture are provided in the supplementary. 
\\
\textbf{Architecture: }The network architecture for all experiments consists of four convolutional blocks, followed by global average pooling and a 
fully connected layer. Each convolutional block is a stack of three convolution layers, each of which is followed by Batch normalization and a ReLU. 
Finally, there exists a pooling layer between two blocks\footnote{More details are provided in the supplementary.}.

\textbf{EXP 1: Confirming the VTSS Hypothesis}

\textbf{Goal:} Recall that the VTSS hypothesis predicts that a VTSS task would learn useful features using a particular transformation $\mathcal{G}$ only when the predicted instantiations of $\mathcal{G}$ do not already exist in data. In this experiment, we test this hypothesis for the VTSS tasks based on rotation \cite{gidaris2018unsupervised} and translation. The overall approach for this experiment is to break the assumption of the VTSS hypothesis \emph{i.e.} the assumption that instantiations from $\mathcal{G}$ are \textit{not} present in the original data or $\mathcal{H}$.  We do this by introducing increasing elements from $\mathcal{G}$ in the original data itself, \textit{independent} of the fact that  the VTSS task would additionally apply and predict instantiations of $\mathcal{G}$ to learn a useful representation. This artificially increases $| \mathcal{G}\cap \mathcal{H} |$. Checking if $C$ (semi-supervised performance of the learned representations) varies inversely allows one to confirm whether the VTSS hypothesis holds.

%Such a self-supervision task for rotation for instance, would be to rotate the input image by $\{ 0, 90^\circ, 180, 270. \}$ and then predicting the augmentation between these four classes. However, to simulate the existence of instantiations of the predicted transformation, we randomly add them into the data \textit{before and independently} of the self-supervision task. This simulates the setting where the instantiations of the predicted transformations already \textit{exist} in the data. 

\textbf{Experimental Setup:} We explore two VTSS tasks based on rotation and translation respectively. The \textit{prediction} range of these transformations are as follows\footnote{We provide more details in the supplementary.}: 

1) \textbf{VTSS Rotation: \cite{gidaris2018unsupervised}} Image rotations by $\{0^\circ, 90^\circ, 180^\circ, 270^\circ\}$ leading to 4-way classification. The input image was rotated by one of the four angles. The VTSS task  was to predict the correct rotation angle applied. This is essentially the same VTSS task employed by RotNet.

2) \textbf{VTSS Translation:} Image translations by 5 pixels with the directions $\{$up, down, left, right, no translation (center)$\}$ leading to a 5-way classification task. From the original image, a center crop with a 5 pixel margin was cropped, which was now considered to be the `no translation' input (center crop). Translations by 5 pixels were applied to this center patch in one of the directions between up, down, left and right. The VTSS task was  to predict which direction the image was translated. The 5 pixel margin allows for a 5 pixel translation with no artifacts. This task based on translations is novel and is part of our contribution.

%3) \textbf{VTSS Scale:} Image scaling up by $\{$0 pix, 4 pix$\}$ leading to a 2-way classification task. A 4 pixel scaling means that the input original image was zoomed in by 4 pixels. This was achieved by cropping a center crop with a margin of 4 pixels, and then resizing that crop to the original image size. The VTSS task was to predict by how much was the input image scaled (0 or 4 pixels). 

\begin{figure}%{r}{0.5\textwidth}
    \begin{center}
        \subfigure[CIFAR 10]{%
        \centering
            \includegraphics[width=0.5\columnwidth,valign=m]{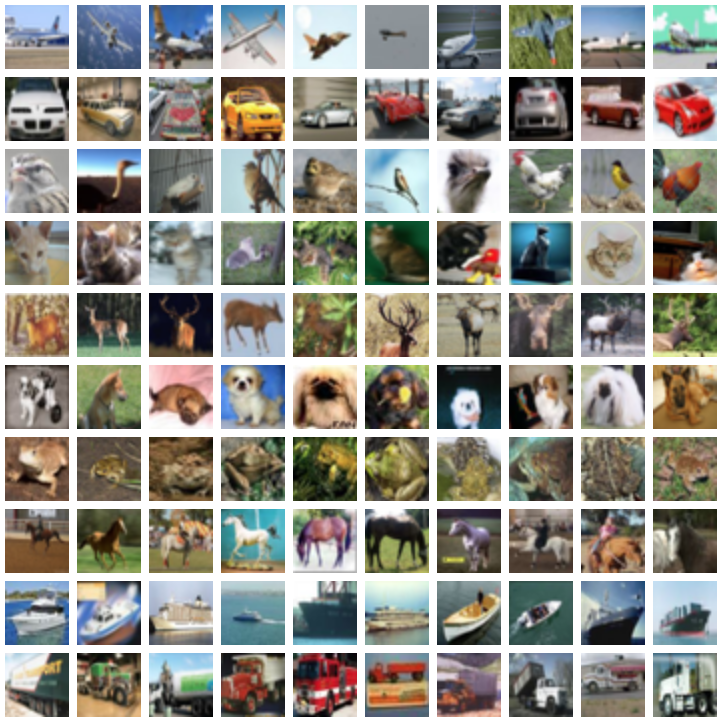}\label{fig_cifar}
        }%
         \subfigure[SVHN]{%
        \centering
        \includegraphics[width=0.5\columnwidth,valign=m]{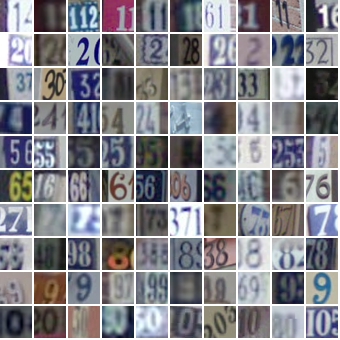}\label{fig_svhn}
        }
        
    \end{center}
    \vspace{-0.5cm}
\caption{  Representative samples from the CIFAR 10 and SVHN dataset. }

    \vspace{-0.5cm}

\end{figure}

%\begin{table}
%%\parbox{.5\linewidth}{
%\small
%\centering
%\begin{tabular}{ | c |  c  | c  | c  |  c  |} % centered columns (4 columns)
% %inserts double horizontal lines
%\hline
%VTSS    &  C10  &   SVHN   & FMNIST  & C100 \\
%\hline
%\hline
%Translation   &   Low   &   Low    &    Low   & Low       \\
%Rotation   &  Low    &   Low    &   Low    &   Low     \\
%Scale   &   High   &    High   &  Low     &      High  \\
%\hline
%%inserts single line
%\end{tabular}
%%\caption{ \textbf{Visual Transformations present in Datasets.} A table summarizing a rough amount of pre-existing transformations in the original data of each set. For rotation, the table denotes the presence of rotations only larger than $90^\circ$. The amount of transformations were estimated only qualitatively through visual inspection. }
%\label{tab_exp_transpresent}
%\vspace{-0.5cm}
%\end{table}

For each transformation $\mathcal{G}$, more instantiations of $\mathcal{G}$ were sequentially added in in the \textit{original data} independent of the corresponding VTSS task. Therefore for each image, there are in fact two separate stages where a transformation is added a) the proposed ablation study itself and b) the VTSS task independent of the ablation study. We now explain the protocol in detail for rotation which has 4 runs (experiments). \textbf{Run 1) Baseline.} $\mathbf{0^\circ}$ The original data contains no rotations added in. This is used for the standard RotNet VTSS task of predicting a 4-way task after rotating the image by one of $\{0^\circ, 90^\circ, 180^\circ, 270^\circ\}$ rotations. This model is evaluated for semi-supervision accuracy and is set as the baseline. \textbf{Run 2)} $\mathbf{0^\circ, 90^\circ}$.  Next, the same procedure is followed however, the \textit{original data} that is sent to the VTSS task, \textit{already} contains all images at $0^\circ$ \textit{and} $90^\circ$ rotations. It is crucial to note however, that the VTSS task of rotating one of $\{0^\circ, 90^\circ, 180^\circ, 270^\circ\}$ rotations and then predicting the rotation remains unchanged. The VTSS task then transforms and predicts based on the original $0^\circ$ images and the images at $90^\circ$ identically. \textbf{Run 3)} $\mathbf{0^\circ, 90^\circ, 180^\circ}$ Now, the same  procedure (VTSS task followed by semi-supervision evaluation) is followed by having all images rotated at each of $\{0^\circ, 90^\circ, 180^\circ\}$ \textbf{Run 4)} $\mathbf{0^\circ, 90^\circ, 180^\circ, 270^\circ}$ Finally, yet another run uses all four rotations added in to all images of the original data. This protocol is followed similarly for translation (predicting 5-way between no translation, up, down, left and right) where the particular transformations that were measured by the VTSS task were added into the original data sequentially. Further details are provided in the supplementary. The performance metric considered for each transformation $\mathcal{G}$ is the semi-supervised accuracy (obtained using the protocol explained in the general experimental settings) on the CIFAR 10 and SVHN test sets. 

\textbf{Results and Discussion.} Table.~\ref{tab_exp_VTSS} showcases the results of these experiments. The left column indicates which transformation instantiations (each for rotation and translation) were added into the original data as part of this ablation study. The semi-supervised accuracy indicates the performance of the learned features towards the downstream classification task (protocol introduced in \cite{gidaris2018unsupervised}). The number in the bracket is the self-supervised accuracy which indicates the test accuracy on the VTSS task itself. Higher accuracy indicates the model is able to distinguish between the between transformations added in. We make a few observations.

\textbf{Observation 1:} We find that VTSS Rotation performs well when there are no rotations already present in the data (both for CIFAR 10 and SVHN). However, the method completely breaks down for both datasets when \textit{all} rotations are present. This indicates that the model has learnt noisy features. 

\textbf{Observation 2:} Notice that the self-supervision classification accuracy for all transformations steadily decreases as more rotations were added into the original data. This is in alignment with Eq.~\ref{eq_vtss_hyp}.  Indeed, as more ablation transformations are added in the original data, it becomes difficult for the VTSS task to learn useful features due to the transformation conflict effect. Observation 1 and 2 together confirm the VTSS hypothesis.

\textbf{Pre-existing Transformations in SVHN and CIFAR:} For the next  observation we take a look at the SVHN  \cite{netzer2011reading} and CIFAR 10 datasets illustrated with a few samples in Fig.~\ref{fig_svhn} and Fig.~\ref{fig_cifar}. For SVHN, we find that there exist considerable scale variation and blur within each digit class.  This blur also acts as scale variation as it simulates the process when a small low resolution object is scaled up leading to blur. However, note that since the dataset was created by extending each digit bounding box in the appropriate directions leading to a square, each digit of interest is almost exactly centered. Thus, there pre-exists very little translation in the dataset. Coupled with the fact that digits have lesser variation than general objects, the visual concepts of interest are more centered. CIFAR 10 on the other hand has more complicated objects also with some scale variation already present in the vanilla dataset. The complex nature of the visual classes results in relatively more translation jitter of visual concepts of interest than SVHN. Lastly, both datasets have some rotation variation however not as extreme as $90^\circ$ or beyond.

% Recall, that the VTSS hypothesis predicts that for a particular VTSS task, if the transformation corresponding to that task pre-exists in a particular dataset, then performance suffers

\textbf{Observation 3:} We observe that VTSS Translation performs closer to the fully supervised performance for SVHN compared to CIFAR 10. Keeping in mind that SVHN has relatively less translation than CIFAR 10, this is consistent with and supports  the VTSS hypothesis.

%\textbf{Observation 4:} We observe that VTSS Translation 

%We find that even the baseline experiment for VTSS Scale fails completely on SVHN. 

%%%%%%%%%%%%%%%%%%%%%%%%%%%%%%%%%%%%%%%%%%%%%
\begin{table}
%\parbox{.5\linewidth}{
\small
\centering
\begin{tabular}{ | c |  c  | c  | c  |  c  |} % centered columns (4 columns)
 %inserts double horizontal lines
\hline
VTSS    &  C10  &   SVHN   & FMNIST  & C100 \\
\hline
\hline
R \cite{gidaris2018unsupervised}   &    89.15  &   91.29      &  \textit{91.94}   &   63.62      \\
T   &   86.20   &    91.16       &     88.98      &      57.10    \\
 S &    43.89  &    28.42       &       83.48       &      17.72   \\
\hline
 R+T              &   \textbf{89.58}  &    \textit{ 91.56  }   &          \textbf{92.18}   &    \textit{64.79 }    \\
S+T               &   71.53   &    87.56      &   88.02       &     45.09     \\
R+S               &   89.00   &       89.16     &     91.81     &   63.78       \\
R+T+S            &   \textit{89.39}  &     \textbf{ 91.72 }    &    91.63     &     \textbf{64.87}   \\

%\hline
% R$\times$T              &      &           \\
%S$\times$T               &      &       &         \\
%R$\times$S               &      &       &         \\
%R$\times$T$\times$S            &      &       &         \\

\hline
FS    3 blocks      &   89.96  &      91.43      &    77.37      &     64.93       \\
FS   4 blocks      &  90.26   &           92.50    &     92.21   &   65.95      \\

\hline
\hline
 S (full) &  31.16    &    7.76     &        87.87  &       15.01
 \\
R (full) &    88.50  &   90.08      &      93.70     &   61.04    \\

FS    3 blocks   (full)   &  91.35   &    80.83        &     92.65     &  52.60       \\
FS 4 blocks   (full)   &    91.66   &    90.57          &     94.62    &  67.95       \\

\hline

%inserts single line
\end{tabular}
\caption{ \textbf{Visual Transformation based Self-Supervision (VTSS) through a combination of transformations.} The column on the left denoted the transformation that the self-supervised backbone was trained with. In the full crop (full) setting, the entire image was utilized for training and testing. In the base setting, the center crop of the image with a margin of 5 pixels on all sides was used. This was done for a better comparison with VTSS Translation which required a 5 pixel margin to allow room for translations as the VTSS task. \textbf{bold} and \textit{italics} indicate the best and second-best performances respectively. VTSS tasks using a combination of transformations performed the best for all four datasets.}
\label{tab_exp_combination_supp}
\vspace{-0.5cm}
\end{table}

%scale: svhn was developed by cropping detection boxes, hence not much jitter exists, but significant scale already

%\textbf{Rotation:} \cite{gidaris2018unsupervised}
% introduced predicting rotations as a method for self-supervision. However, revisiting the algorithm from the perspective of the VTSS hypothesis, we investigate that one of the primary reasons the method is effective is the fact that the dataset does not have all predicted instatniations of rotation.  

\textbf{EXP 2: Exploring VTSS Tasks with Multiple Transformations Simultaneously}

\textbf{Goal:} Though self-supervision with images has shown considerable promise as an unsupervised technique, it is still a fairly recent paradigm. Typically, self supervision using visual transformations has been applied with a single transformation type, for instance exclusively rotations for RotNet \cite{gidaris2018unsupervised}. Given that in this study, we have demonstrated the existence of a VTSS technique for translation and scale as well, it is natural to ask the question: \textit{how does the performance differ when using multiple transformations in conjunction?}. We explore answers and also observe phenomenon that the VTSS hypothesis predicts.

\textbf{Experimental Setup:} For the datasets CIFAR 10, 100, FMNIST and SVHN, we train the standard backbone network with 4 convolution blocks with VTSS tasks of Rotation $\{0^\circ, 90^\circ, 180^\circ, 270^\circ\}$, Translation $\{$up, down, left, right, no translation (center)$\}$ with a shift of 5 pixels and Scale $\{$0 pix, 2 pix zoom, 4 pix zoom$\}$. However, VTSS Translation needs a small margin (to translate without artifacts). We apply this crop margin (of 5 pixels) to all data for all tasks. Therefore, the default images for all tasks is the 5 pixel margin center crop of the original image. For VTSS Scale, this crop was designated to be the 0 pix zoom. A 2 pix zoom (or 4 pix zoom) would perform yet another crop with a 2 pix (or 4 pix) margin on each side before resizing the image back to the center crop size.  We combine two or more transformations in an additive fashion. For instance, if VTSS Rotation predicts 4 classes and VTSS Translation predicts 5 classes, the task VTSS Rotation + Translation will predict 4+5-1 = 8 classes overall (where we combine the identity class of all transformations into a single class). We perform experiments with all 4 combinations between the three transformations. Additionally, we run each transformation individually to serve as a baseline under the center crop setting. All individual and combination transformations were run in the center crop setting to be consistent in data size for the VTSS Translation task for the combination experiments. However, in practice, when VTSS Scale and Rotation \cite{gidaris2018unsupervised} would be applied independently, the entire image would be used and not just the center crop. Thus we provide additional results with just the individuals transformations of VTSS Rotation and Scale on the full sized crop of side 32 (without any center cropping). The corresponding fully-supervised results with the full crop were also provided.

%\textbf{ Additional Runs using Full Crop for VTSS Rotation and Scale:} In the main paper, the corresponding table (Table 2, in the main paper) showcased results only with all images cropped with a center-crop of a 5 pixel margin. This was done in order to be consistent in data size for the VTSS Translation task for the combination experiments. However, in practice, when VTSS Scale and Rotation \cite{gidaris2018unsupervised} would be applied independently, the entire image would be used and not just the center crop. Thus we provide additional results with just the individuals transformations of VTSS Rotation and Scale on the full sized crop of side 32 (without any center cropping). The corresponding fully-supervised results with the full crop were also provided.

\begin{figure}%{r}{0.5\textwidth}
    \begin{center}

        \includegraphics[width=0.95\columnwidth,valign=m]{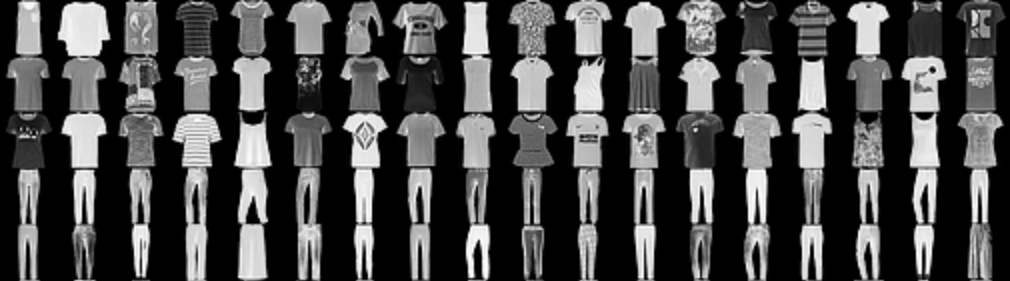}

    \end{center}
    %\vspace{-0.5cm}
\caption{ Samples from the Fashion-MNIST dataset. Note that compared to CIFAR 10 and SVHN, FMNIST contains considerably less scale variation. Thus, the VTSS hypothesis predicts that VTSS Scale would perform better, than that of CIFAR 10 and SVHN, which is indeed the case from Table.~\ref{tab_exp_combination_supp}. This is yet another confirmation of the VTSS hypothesis. }
\label{fig_fmnist}
    \vspace{-0.5cm}

\end{figure}

\textbf{Pre-existing Transformations in Fashion-MNIST:} A few sample images from the Fashion-MNIST dataset are shown in Fig.~\ref{fig_fmnist}. One notices immediately that the dataset contains little to no translation jitter, no rotation and importantly, very little scale variation as compared to CIFAR and SVHN (see corresponding figure in main paper). This implies that the VTSS hypothesis would predict that VTSS Scale would be effective. The FMNIST dataset hence is a good dataset to prove  effectiveness of the VTSS Scale task. The dataset contains 60,000 training images and 10,000 testing images. each image is sized 28, which for our experiments was rescaled to 32. This does not affect overall trends in our experiments since all images were resized equally.

%\textbf{Results and Discussion:} The results of these experiments are presented in Table.~\ref{tab_exp_combination_supp}.

\textbf{Results: Individual Transformations.} The results of these experiments are presented in Table.~\ref{tab_exp_combination_supp}. We find that VTSS Rotation overall performs consistently high. However, given that SVHN has lesser translation (see discussion on pre-existing transformations in datasets), VTSS Translation performs better on SVHN than CIFAR 10 and 100. This is indeed consistent with the VTSS hypothesis. Note that scale performs worse on both the CIFAR datasets and SVHN. Recalling the prior discussion regarding the presence of scale variation in both CIFARs and SVHN, this result is consistent with the VTSS hypothesis. In fact, due to the presence of more blur which acts as scale variation, VTSS Scale works worse on SVHN than CIFAR 10, which has no common blurry artifacts. This observation as well is consistent with the VTSS hypothesis.  Interestingly however, that given the observation that FMNIST has considerably less scale variation than CIFAR and SVHN, the VTSS hypothesis predicts that VTSS Scale would perform better on FMNIST than on CIFAR and SVHN. Indeed, this is what we observe. Both the full crop and the center crop VTSS Scale performance on FMNIST are significantly higher than that of CIFAR and SVHN. This provides further evidence towards the confirmation of the VTSS hypothesis.

\textbf{Results: Combinations of Transformations.} We find that a combination of VTSS R + T works better than isolated VTSS Rotation for all four datasets. This also true for VTSS R+T+S for all datasets except FMNIST. This is the first evidence that utilizing multiple transformations simultaneously as a single VTSS task can provide performance gains over any individual transformation. Given that it is computationally inexpensive to train under such a setting, this result is encouraging. Notice also that even though VTSS Scale performs worse on SVHN than CIFAR, the combination VTSS S+T performs better on CIFAR. Nonetheless, due to the inherent presence of scale in CIFAR and SVHN, the VTSS hypothesis predicts that VTSS tasks involving scale would suffer in performance. However, scale achieves more success on FMNIST due to the absense of inherent scale (a hypothesis prediction). This is something we do observe in Table.~\ref{tab_exp_combination_supp}. From these experiments, we conclude that there is benefit in combining VTSS tasks for different transformations, however it must be done so while being aware of what transformations or factors of variation already exist in the data. Indeed, VTSS tasks using some sort of combination
of transformations consistently outperformed all individual transformations for all four datasets. 

%In that regard, it is atypical to observe in-plane rotations of visual concepts more than $90^\circ$ existing naturally in visual data generated by the real world. The 

%Finally, we find that the combination of all three transformations outperforms all other settings. Interestingly, it outperforms the fully-supervised of 3 conv blocks which has the same number of parameters. This suggests that much is to be gained by the exploration of combination of self-supervision tasks. 

 %We find that the combination of transformation experiments follow similar trends to those of CIFAR and SVHN. VTSS R+T outperforms all other combinations. We however find that, VTSS Rotation and Translation perform slightly worse when combined with Scale. 

\textbf{Overall Observation: Effectiveness of VTSS Rotation} Our results indicate that VTSS Rotation seems to consistently perform better than translation and scale when applied individually. The VTSS hypothesis also begins to offer a probable justification as to why. The degree to which translation and scale are applied as part of the VTSS task to learn these features (which were effective nonetheless), are small. Larger variation we found typically reduced performance given a fixed sized image (see Fig.~1(b) and Fig.~1(c) in the supplementary). These transformations are likely to exist in subtle amounts at similar ranges to those that were applied as part of the VTSS task. This leads to the detrimental effect of transformation conflict (see Fig.~\ref{fig_transformationconflict}). On the other hand, VTSS Rotation  was found to work well at  large ranges ($90^\circ$) in the original study \cite{gidaris2018unsupervised}. Nonetheless, rotation seldom occurs naturally in most datasets at such large ranges (including real-world ones). \textit{Our VTSS hypothesis therefore predicts that rotation is a particularly well-suited transformation for VTSS for general visual data}. 
%(see Fig.~\ref{fig_trans_range} and Fig.~\ref{fig_scale_range})
%In light of these observations, VTSS Rotation \cite{gidaris2018unsupervised} indeed has an advantage and is expected to be most effective atleast among the studied visual transformation based methods.

%\textbf{Data Preprocessing:}

%explain why very novel
%explain that cifar translation works worse because svhn qualitatively seems to have lesser translation invariance than cifar
%SVHN has considerable scale variation \cite{goodfellow2013multi} (section 5.1)

\textbf{Concluding Remarks.}  Unsupervised learning of representations is a keystone in the path to general intelligence. Indeed, learning representations through self-supervision has emerged as a promising way forward addressing a multitude of problems in machine learning simultaneously. Despite the plethora of self-supervised techniques that have emerged as effective, our fundamental understanding both theoretical and empirical is still limited. This study hopes to provide more insight into this phenomenon and to serve as a useful reference in the development of better empirical techniques and more rigorous theory within the field of self-supervision.

\newpage
\section{Appendix: Additional Ablation Studies, Details and Observations}

%We validate our hypothesis through experiments on CIFAR 10 using multiple mainstream algorithms that were previously investigated such as RotNet \cite{gidaris2018unsupervised,hendrycks2019using}, Colorization and Jigsaw. 

\textbf{General Hyperparameters.} For each transformation and dataset, the evaluation protocol and hyperparameters remained constant. The  self-supervised  backbone network and the fully-supervised network were both trained on the training set for all training samples (unless specified) for 200 epochs.  The  learning rate was set at 0.1 and multiplied by 0.02 at 60, 120 and 180 epochs with SGD with a batch-size of 128, momentum of 0.9 and weight decay of $5\times 10^{-4}$. However, for every transformation to added in for a particular VTSS task, \emph{e.g.} rotation, the entire batch was transformed by that augmentation and added to the batch. However, for the VTSS hypothesis confirmation studies only the transformations were added into this original 128 sized batch as an ablation study. For this, if there are $K$ different instantiations of the transformations to be added in, then the 128 sized batch was divided by $K$ and each shard was transformed by one instantiation. Once the self-supervised network was trained, the weights till the second conv block were frozen and a classifier on top was added. The tables in the main paper (Tables 1 and 2) were not performed with any data augmentation. However,for the ablation studies in this supplementary, we performed standard data augmentation of random crops with a 2 pixel padding with randomized horizontal flips. Interestingly, we find that VTSS translation is effective even with such augmentation.

\begin{figure*}%{r}{0.5\textwidth}
    \begin{center}
        \subfigure[Rotation]{%
        \centering
            \includegraphics[width=0.65\columnwidth,valign=m]{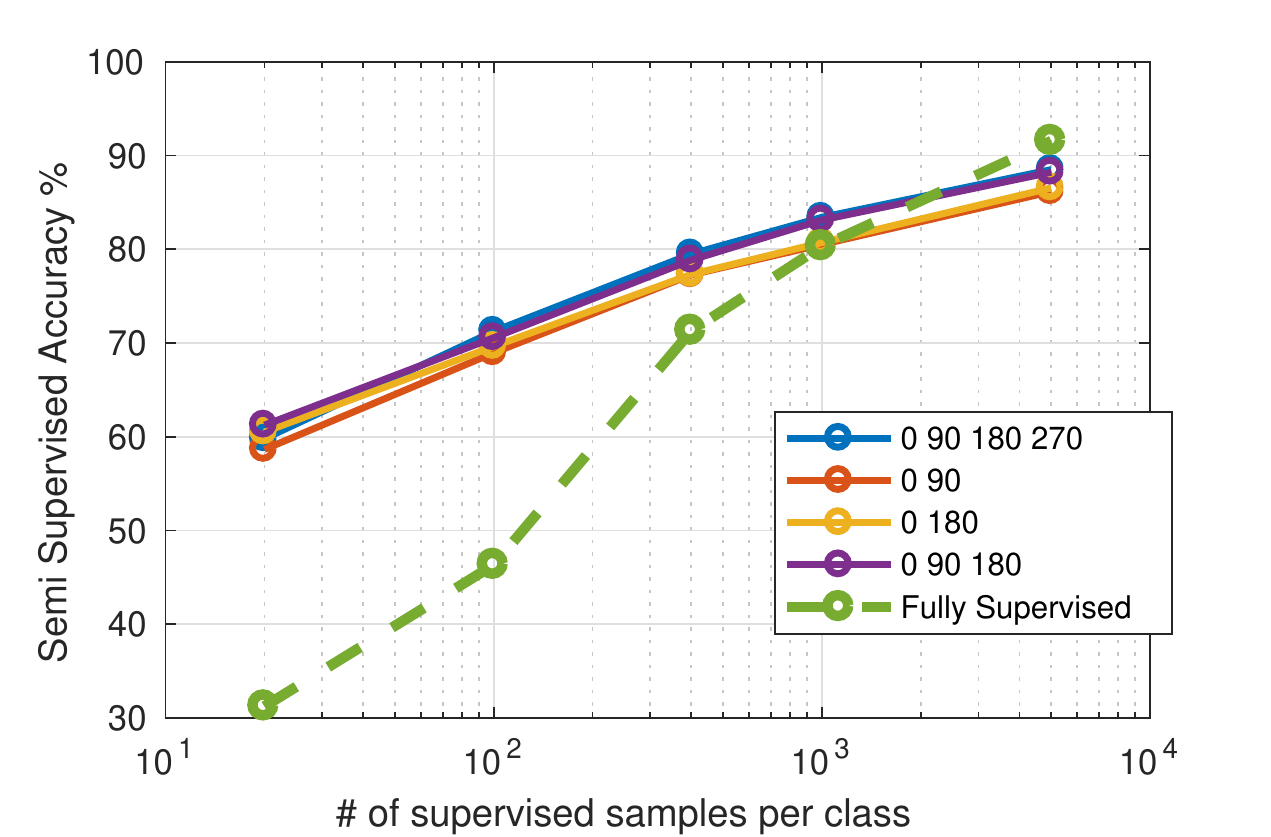}\label{fig_rot_range}

        }%
         \subfigure[Translation]{%
        \centering
        \includegraphics[width=0.65\columnwidth,valign=m]{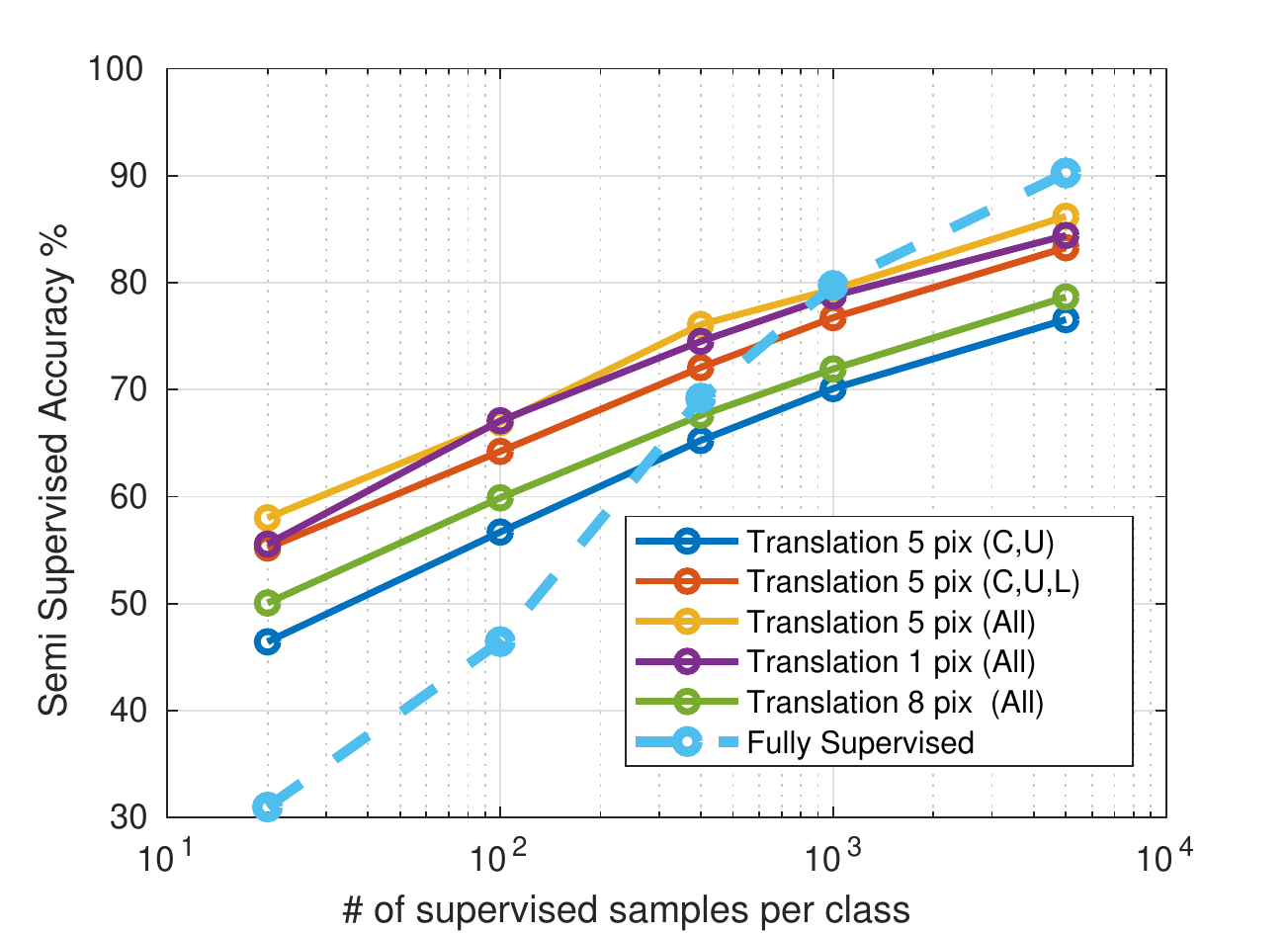}\label{fig_trans_range}
        }
         \subfigure[Scale]{%
        \centering
        \includegraphics[width=0.65\columnwidth,valign=m]{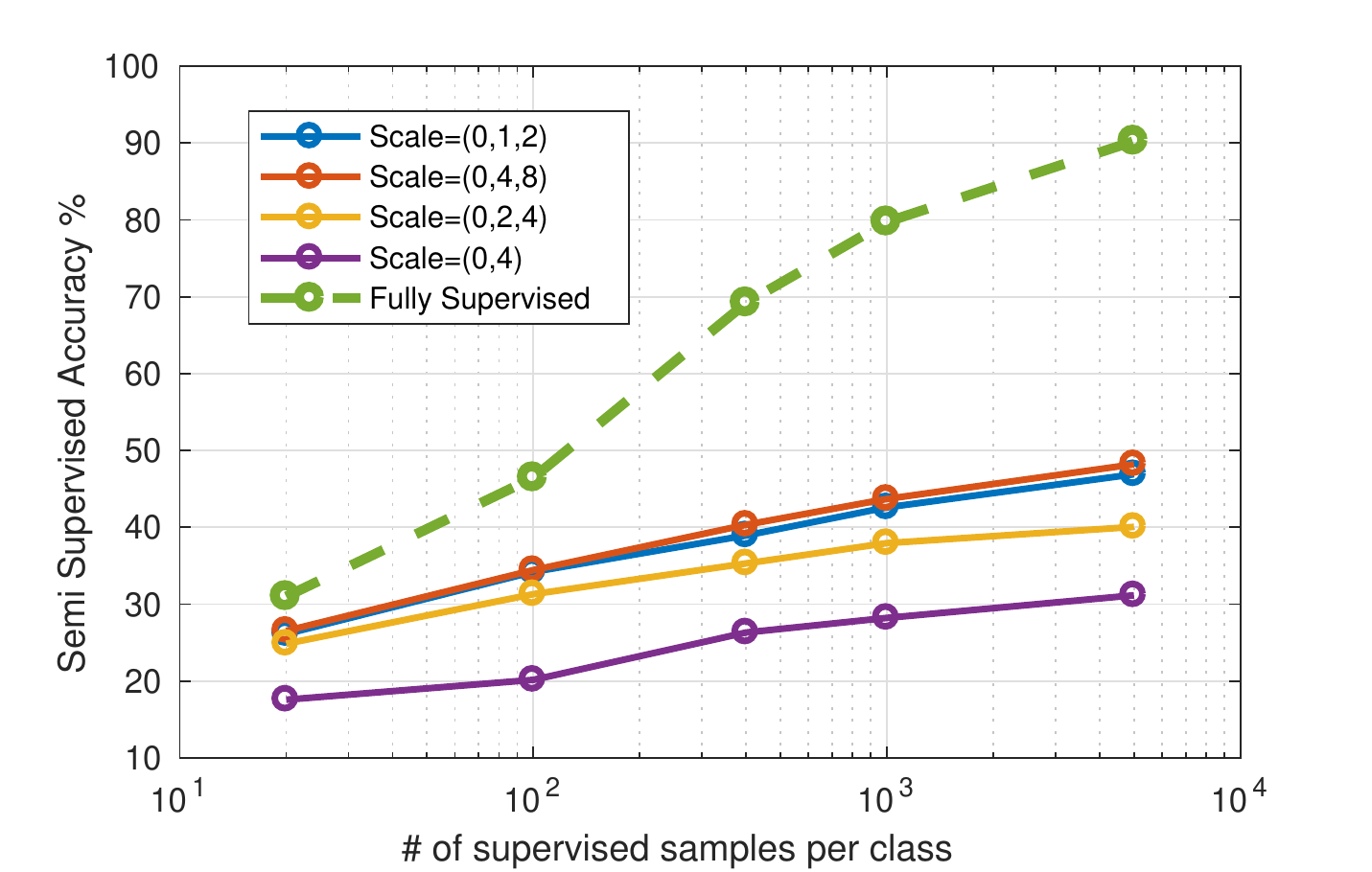}\label{fig_scale_range}

        }
        
    \end{center}
    \vspace{-0.5cm}
\caption{ \textbf{Results of the Ablation study:} Effect of Transformation Range on VTSS Rotation \cite{gidaris2018unsupervised},Translation and Scale on CIFAR 10.   }
    \vspace{-0.5cm}

\end{figure*}

\textbf{Further Details in Architecture.} All networks consisted of units or blocks called a conv block. Each conv block consisted of 3 conv layers with 192 channels, each followed by batch normalization and ReLU. The fully-supervised and self-supervised backbone networks consisted of 4 conv blocks (unless specified otherwise) with average pooling of kernel size 3, stride 2 and padding 1 after each block. The semi-supervised classifier was trained on conv block 2 features after training the self-supervision model. The semi-supervised classifier added consisted of a single conv block with 192 channels, global average pooling followed by a single linear layer. 

%\textbf{Architecture: }The network architecture for all experiments consists of four convolutional blocks, followed by global average pooling and a fully connected layer. Each convolutional block is a stack of three Convolution layers, each of which is followed by Batch normalization and a ReLU. Finally, each there exists a pooling layer between two blocks\footnote{More details are provided in the supplementary.}.

 \textbf{Ablation A: Effect of Transformation Range.} We trained a backbone feature extractor network (RotNet) with the VTSS rotation task for different sets of rotations\footnote{We provide additional ablation studies in the supplementary} on CIFAR 10.  There was no independent rotations added (as in our VTSS hypothesis confirmation study) other than the rotations added by the VTSS task itself. We also perform this experiment similarly for the VTSS translation and scale tasks. In this case, we steadily increase the number of directions the translation is added in and correspondingly increase the number of classes for prediction. Lastly, the pixel range of the translation was also varied. For VTSS Scale, the number of scales and the range was varied.

\textbf{Results.} The results of this experiment are presented in Fig.~\ref{fig_rot_range},  Fig.~\ref{fig_trans_range} and Fig.~\ref{fig_scale_range}. We find that though predicting more rotation angles help, the improvement is marginal. Further, predicting even a single instantiation of the transformation resulted in the learning of useful representations. These results are consistent with those reported in the original study \cite{gidaris2018unsupervised}. In this case of translation however, we find a significant increase initially after which there are diminishing returns.  It is interesting to note that VTSS translation learnt useful features even with a single pixel shift (on each side) (while using traditional data augmentation). In fact, the 5 pixel shift performs only marginally better. However, a 8 pixel shift (therefore a total crop of just 32-16=16) deems excessive for a 32 sized dataset and drastically decreases performance.  For VTSS Scale, it was difficult to find an overall trend. Nonetheless, we find that representations learned are in general poor. This we hypothesis is largely due to the existence of scale variation already in CIFAR. However, it is interesting to note that a scale variation of even 1 and 2 pixels can be useful for representation learning.

\textbf{Ablation B: Effect of number of self-supervised and supervised samples.} Typically, self supervision tasks are trained on as much data as possible. This is primarily due to the availability of inexpensive self-labels. However, we explore the case where there is a imbalance of data between the self-supervision task and semi-supervision tasks. We increase the number of samples available per class through $\{20, 100, 400, 1000, 5000\}$ samples for both the VTSS Rotation and the downstream semi-supervision tasks. 

\begin{figure}%{r}{0.5\textwidth}
    \begin{center}

        \includegraphics[width=0.85\columnwidth,valign=m]{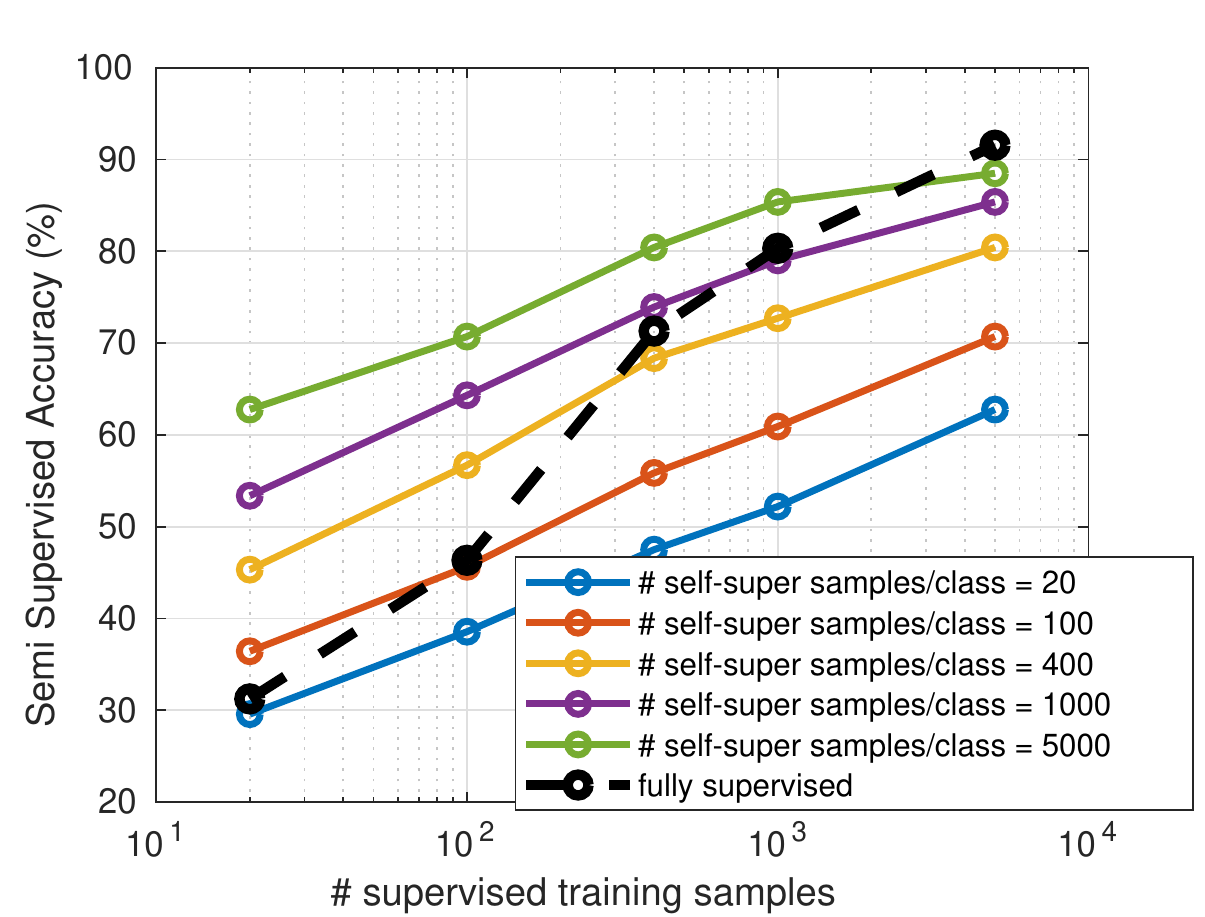}

    \end{center}
    %\vspace{-0.5cm}
\caption{ \textbf{Results of Ablation B.} Effect of number of self-supervised and
supervised samples.}
\label{fig_exp_numsamples}
    \vspace{-0.2cm}

\end{figure}

\textbf{Ablation B: Results.} Fig.~\ref{fig_exp_numsamples} showcases the results of this experiment. Interestingly, we find that the performance increases linearly at almost identical rates for both the self-supervision and the downstream semi-supervision tasks. For instance, the performance of a 1000 samples/class for VTSS and just 20 samples/class for semi-supervised learning is very similar to 20 samples/class for VTSS and 1000 samples/class for semi-supervised learning. We find similar trends for other settings. This highlights the benefits of VTSS tasks. For a  particular amount of data with an inexpensive self-labelling scheme, VTSS provides a level of performance to the downstream classifier similar to that of a semi-supervised model which was trained with the same amount of \textit{labeled} data. Nonetheless, such linear parallels between VTSS and downstream semi-supervision are encouraging.

\textbf{Ablation C: Effect of number of classes used for self-supervision.} VTSS tasks are typically applied to a wide array of data. However, what is the level of returns that a VTSS task provides given a steady increase in both diversity and amount of data? For this experiment, we train the same 4 conv block layered network with VTSS Rotation given $\{1, 2, 3, 5, 8, 10  \}$ classes. We are interested in the trends with which the performance increases w.r.t the downstream semi-supervised accuracy.

\begin{figure}%{r}{0.5\textwidth}
    \begin{center}
        \includegraphics[width=0.85\columnwidth,valign=m]{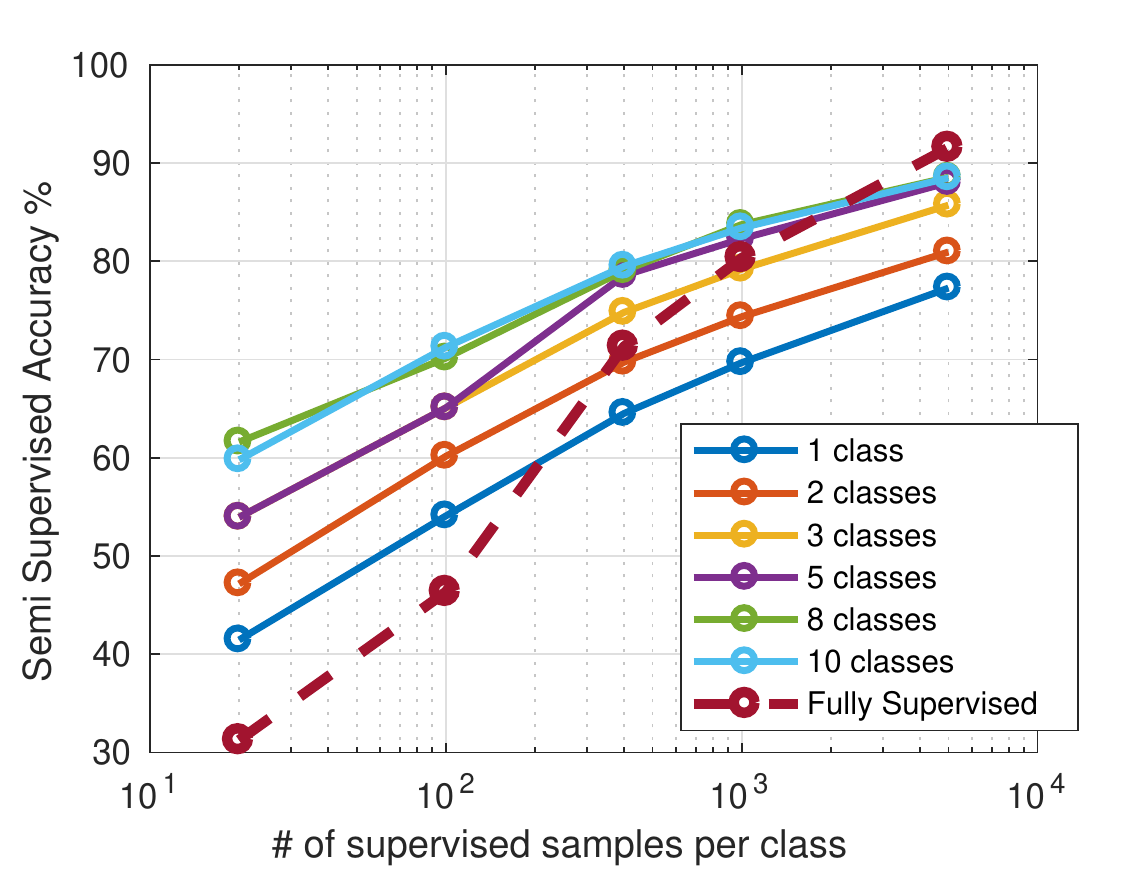}
    \end{center}
    %\vspace{-0.5cm}
\caption{ \textbf{Results of Ablation C.}  Effect of number of classes used for self-supervision.}
\label{fig_exp_num_classes}
    \vspace{-0.5cm}%

\end{figure}

\textbf{Ablation C: Results.} The results of this experiment are presented in Fi.g~\ref{fig_exp_num_classes}. We find that the downstream semi-supervision performance increases as expected. However, the returns are diminishing. Indeed, though when there are 5000 samples/class available for training the semi-supervised network, the performance saturates with just 5 classes=5000 samples in total for the VTSS task. Similar trends are observed for the cases where there are lower samples/class available for semi-supervised learning. This suggests that though VTSS tasks are powerful, they seem to be hitting a barrier to the diversity of features that the model can learn. Attention must be paid to other aspects of the learning problem such as the size of the network, architecture etc. \cite{kolesnikov2019revisiting} in order to further allow improvements leveraging more self-supervised data.

\textbf{Additional Observation: Effectiveness of VTSS tasks in general.} Taking a step back from a detailed inter-transformation analysis, we observe the performance of self-supervision followed by semi-supervised tasks. Recall that the original VTSS backbone network consisted of 4 conv blocks of three conv layers each. Then, only the first two conv blocks were used as a fixed feature extractor for the semi-supervised classifier on top that was trained. This classifier consisted of a single conv block that is identical to the other blocks. Therefore, in Table 1 and Table 2 in the main paper, the fully supervised networks or FS with 3 blocks had similar complexity to the overall self and then semi-supervised models, and provided a fair comparison from the perspective of model complexity. Yet, in the case of SVHN and FMNIST, we find that VTSS Rotation performs better than FS 3 blocks for the full crop setting (including scale for FMNIST). For the center crop experiments, VTSS R+T performs better than FS 3 blocks for SVHN. This showcases the overall effectiveness of VTSS tasks in general compared to fully-supervised networks of similar complexity.

{\small
\bibliographystyle{icml2020}
\bibliography{example_paper}
}

\end{document}